# Certain Bayesian Network based on Fuzzy knowledge Bases

Abdelkader HENI [1,2] , abdelkader.heni@edunet.tn
Mohamed-Nazih OMRI[1,2] , nazih.omri@ipeim.rnu.tn
Adel M ALIMI, [1,3] , adel.alimi@enis.rnu.tn

[1] REGIM: Research Group on Intelligent Machines
[2] Department of Technology-Preparatory Institute of Engineering Studies
5019 Monastir, Tunisia
[3] National School of Engineers
*3000 Sfax, Tunisia*

## Abstract

*In this paper, we are trying to examine trade offs between fuzzy logic and certain Bayesian networks and we propose to combine their respective advantages into fuzzy certain Bayesian networks (FCBN), a certain Bayesian networks of fuzzy random variables. This paper deals with different definitions and classifications of uncertainty, sources of uncertainty, and theories and methodologies presented to deal with uncertainty. Fuzzification of crisp certainty degrees to fuzzy variables improves the quality of the network and tends to bring smoothness and robustness in the network performance. The aim is to provide a new approach for decision under uncertainty that combines three methodologies: Bayesian networks certainty distribution and fuzzy logic.*
*Within the framework proposed in this paper, we address the issue of extending the certain networks to a fuzzy certain networks in order to cope with a vagueness and limitations of existing models for decision under imprecise and uncertain knowledge*

## 1. Introduction

In this paper, a new knowledge representing model called certain Bayesian network with fuzzy knowledge bases is proposed to cope with uncertain information. We propose the use of fuzzy necessity (i.e. certainty) values for representing uncertain and imprecise information. The fuzzy necessity value is adopted for its capability to express the "possibility" of the degree of certainty of a fuzzy proposition.

We propose a mechanism for defining fuzzy propositions with fuzzy certainty values. There are three steps involved. First, the certain Bayesian network definition with prior fuzzy certainties values, second fuzzy certain nodes and values are transformed into a set of uncertain classical propositions with fuzzy necessities to construct another structure called local knowledge base defined in each standard form in [9] and [10] by means of tow rules. Third, we reverse the process in the second step to synthesize the results obtained in this step into a certain Bayesian network.

This work proposes a novel approach for modelling problems with Bayesian networks involving fuzzy certainty variables. The proposed method formulates reasoning problems using the combination of the three approaches cited above. The model developed here can be built on any exact propagation methods, including clustering, joint tree decomposition, etc.

The organization of this paper is as follows. A description of the representation process of uncertain imprecise propositions and its semantics are defined in next section. An overview of basic concepts is proposed in the section 3. In Section 4, 5 and 6 our approach to model uncertainty with certain Bayesian networks and fuzzy knowledge bases is discussed. Many of related work are described in many parts of this paper. Finally, a summary of our approach and its drawbacks are given in conclusion.

## 2. Representing uncertain and imprecise information

Fuzziness and randomness are two distinct components of uncertainty. While fuzzy sets are a rigorous softening of random sets, many of the operations defined in fuzzy logic lack a complete formalism, and are not strongly supported by experimental evidence.

On the other hand causal possibilistic networks (CPN) or possibilistic Bayesian networks provide an ultimately flexible inference mechanism based on certainty or possibility distribution principles. However, CPNs suffer from the overwhelmingly large conditional possibility tables with discrete variables likewise probabilistic networks namely Bayesian networks.

Fuzzification of continuous or crisp variables reduces the size of conditional certainty tables (case of certain networks in witch we use certainty degrees instead of possibilities or probabilities) to practically acceptable levels and these tables exhaustively encompass all the intuitive and fuzzy rules for inference problems.

In this way, we reach a new knowledge representing engine, called fuzzy certain Bayesian networks, which provides a rigorous formalism for inference under fuzziness and randomness.

The main idea of our modeling system of uncertain and imprecise knowledge is inspired from [3] where the authors have extended Dubois and Prade's [2] definition about the possibility and necessity measures of classical propositions to the case of fuzzy propositions through fuzzy truth values [1]. A classical proposition is true in some possible worlds and false in the rest of possible worlds, while a fuzzy proposition $P$ is true with respect to a possible world to a degree [3]. We model our uncertainty about the actual world by defining a fuzzy necessity distribution over all possible worlds to specify the degree of necessity that the actual world is in each certain world.

More formally, we consider that each necessity degree associated to a propositional formula $P$ is a fuzzy number mapping from an interval $[\beta_1, \beta_2] \subseteq [0, 1]$ to $[0, 1]$ where $[\beta_1, \beta_2]$ is given by expert. Here the exact necessity degree is not specified but we are "certain" that it belongs to the interval $[\beta_1, \beta_2]$.

For each state (respectively for each formula), we associate a fuzzy necessity degree $\chi \in [\beta_1, \beta_2]$. And then we map the fuzzy necessity distribution defined by a membership function $\mu$ ($\chi_1$). More formally:

$$\mu : [\beta_1, \beta_2] \longrightarrow [0 \ 1] \qquad (1)$$
$$\chi_1 \longmapsto \mu(\chi_1)$$

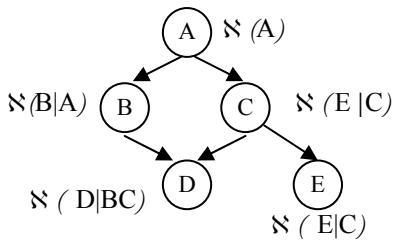

*Fig 1. Fuzzy certain Bayesian network*

Figure1 illustrates the structure of fuzzy certain Bayesian network with five nodes where nodes may receive fuzzy values. $\aleph(*)$ is the associated fuzzy certainty (i.e. necessity) distribution. More explanation of necessity distribution and fuzzy logic will be presented in the second part of next section.

## 3. Preliminary concepts
### 3.1. Fuzzy logic
The fuzzy set theory was first presented by Zadeh [5]. In this theory, uncertainty has more to do with vague definitions of criteria than randomness. The basis of the fuzzy set theory is a membership function $\mu(x)$. This function describes the degree to which a certain statement is true [6]. For example, the statement 'Jhon is tall' can be more or less true.

If we had a precise definition, such as 'Jhon is tall if it has at least 1.75 of size', the membership function would only have values 0 and 1, and fuzzy sets would not be needed. Otherwise, the membership function for a fuzzy set $\hat{A}$ could be:

$$\mu_{\hat{A}} : X \to [0, 1]$$

And $\mu_{\hat{A}}(x)$ is interpreted as the degree of membership of element $x$ in fuzzy set $\hat{A}$ for each $x \in X$.

### 3.2. Possibilistic logic
Possibilistic logic [4] is logic of uncertainty to reason with classical propositions under incomplete information and partially inconsistent knowledge. Formulas of the necessity-valued fragment of possibilistic logic are of the form $(\varphi, \alpha)$ where $\varphi$ is a classical (propositional or first-order) formula and $\alpha \in [0,1]$ is understood as a lower bound for the necessity degree of $\varphi$.

Let $\mathcal{L}$ be a finite propositional language. $p$; $q$; $r$; . . . denote propositional formulae. $\top$ and $\bot$, respectively, denote tautologies and contradictions. $\vdash$ denotes the classical syntactic inference relation. $\Omega$ is the set of classical interpretations $\omega$ of $\mathcal{L}$, and $[p]$ is the set of classical models of $p$ (i.e. interpretations where $p$ is true $\{\omega \mid \omega \models p\}$) [9].

### 3.2.1 Fuzzy Possibility and fuzzy Certainty distributions
The basic element of possibility theory is the possibility distribution $\Pi$ which is a mapping from $\Omega$ to the interval $[0,1]$. Formally a possibility distribution is defined as :

$$\Pi : \Omega \longrightarrow [0 \ 1] \qquad (2)$$
$$\omega \longmapsto \pi(\omega)$$

Where the degree $\pi(\omega)$ represents the compatibility of $\omega$ with the available information (or beliefs) about the real world. By convention, $\pi(\omega)= 0$ means that the interpretation $\omega$ is impossible, and $\pi(\omega) = 1$ means that nothing prevents $\omega$ from being the real world. [7].

Given a possibility distribution $\pi$, two different ways of rank ordering formulae of the language are defined from this possibility distribution. This is obtained using two mappings grading, respectively, the possibility and the certainty of a formula $p$:

· The possibility (or consistency) degree:

$$\prod(p) = max \ ( \ \pi(\omega) : \omega \in [p]) \qquad (3)$$

Which evaluates the extent to which $p$ is consistent with the available beliefs expressed by $[p]$. It satisfies:

$$\forall p, \ \forall q \qquad \prod(p \vee q) = max \ (\prod(p), \prod(q)) \qquad (4)$$

Analogously we can define the fuzzy possibility distribution as follow:
· The fuzzy possibility (or consistency) degree:

$$\mu_{\prod(p)}(t) = max \ (\pi(\omega) : \omega \in [p] \text{ and } \mu_p(\omega) = t) \qquad (5)$$

Where $\mu_p$ denotes the fuzzy set of possible worlds of $p$ in $\Omega$. $\omega$ is a possible world in $\Omega$ and $t$ is the degree of truth. $\mu_{\Pi(p)}$ can be viewed as the possibility measure of a set of possible worlds in which the truth degree of $p$ is equivalent to $t$ [3], i.e.,

$$\forall p, \ \forall q \quad \mu_\Pi (p \vee q) = \max (\mu_\Pi(p), \mu_\Pi(q)) \qquad (6)$$

• The necessity (or certainty, entailment) degree

$$N(p) \ = 1 - \Pi (\neg p) \qquad (7)$$

Which evaluates the extent to which $p$ is entailed by the available beliefs. We have [13]:

$$\forall p, \ \forall q \quad N (p \wedge q) = \min (N (p), N (q)) \qquad (8)$$

Based on [3], we can define our fuzzy necessity as follow:

• The fuzzy necessity (or certainty, entailment) degree:

$$\mu_{N(p)}(t)(p) \ = \ 1 - \mu_{N(p)} (t) (\neg p) \qquad (9)$$

Witch satisfies:

$$\forall p, \ \forall q \quad \mu_N (p \wedge q) = \min (\mu_N(p), \mu_N (q)) \qquad (10)$$

In both systems (certainty distribution and fuzzy certainty distribution) the fuzzy statement: "its almost sure that $\varphi$ is $\omega$ ($\omega \in [\varphi]$)" where $[\varphi]$ denotes the set of models of $\varphi$, can be represented by certainty weighted formula of the form $(\varphi, \alpha)$ where $\varphi$ is a classical propositional formula and $\alpha$ is the lower bound of necessity degree and can be interpreted as a crisp value in case of necessity distribution and as $\mu_N (\chi) \in [\beta_1, \beta_2]$ in the case of fuzzy necessity distribution.

## 4. Certain Knowledge bases: definitions and concepts

A certain formula (i.e. possibilistic formula) is a pair $(\varphi, \alpha)$ where $\varphi$ is a classical first-order closed formula and $\alpha \in [0,1]$ is a positive number. $(\varphi, \alpha)$ expresses that $\varphi$ is certain at least to the degree $\alpha$, i.e. $N(\varphi) \geq \alpha$, where $N$ is a necessity measure modelling our possibly incomplete state of knowledge. The right part of a certain formula, i.e. $\alpha$, is called the *weight* of the formula.

Thus a certain knowledge base $\Sigma$ is defined as the set of weighted formulae. More formally $\Sigma = \{(\varphi_i, \alpha_i), i = 1....m\}$ where $\varphi_i$ is a propositional formula and $\alpha_i$ is the lower bound of necessity accorded to this formula (certainty degree).

### 4.1. Certain Bayesian networks (CBN)

A standard certain Bayesian network is a decomposition of a multivariate necessity distribution according to:

$$N (A_1, ....., A_n) = \min_{i=1..n} N (A_i \mid parents(A_i)) \qquad (11)$$

where $parents(A_i)$ is the set of parents of variable $A_i$, which is made as small as certain by exploiting conditional independencies of the type indicated above [9] and [10]. Such a network is usually represented as a directed graph in which there is an edge from each of the parents to the conditioned variable (fig 2).

In our work a certain Bayesian networks is considered as a graphical representation of uncertain information. It offers an alternative to probabilistic causal network when numerical data are not available.

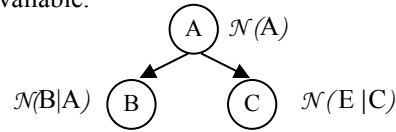

*Fig 2: Example of a certain Bayesian network*

Let $V = \{A_1, A_2,...A_n\}$ a set of variables (i.e attributes or proprieties). The set of interpretations is the Cartesian product of all domains of attributes in $V$. When each attribute is binary, domains are denoted by $D_i = \{a_i, \neg a_i\}$.

A *certain graph* denoted by $NG$ is an acyclic graph where nodes represents attributes i.e a patient temperature and edges represent causal links between them. Uncertainty is represented by necessities distribution, and conditional necessities for each attribute explaining the link force between them.

The conditional necessities distributions are associated to the graph as follow:

For each root attribute $A_i$, we specify the prior necessity distribution $N(a_i)$, $N(\neg a_i)$ with the constraint that:

$$N(a_i) = 1 \quad \vdash \quad N(\neg a_i) = 0 \qquad (12)$$

- For other attributes $A_j$, we specify the conditional necessities distribution $N(a_j | u_j)$, $N(\neg a_j | u_j)$ with $\max(N(a_i | u_j), N(\neg a_i | u_j)) = 1$ where $u_j$ is an instance of $a_j$ parents.

*Example*: the next figure gives an example of certain Bayesian networks with four nodes and their conditional necessities.

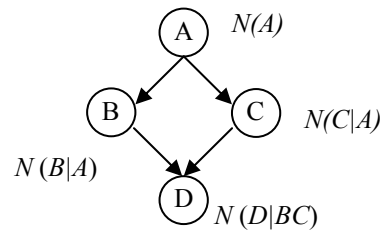

*Fig 3. A certain Bayesian network with four nodes*

The joint certainty distribution is obtained then by applying the chain rule:

$$\mathcal{N}(A_1,...,A_n) = min(N(A_i|U(A_i)) \qquad (13)$$

Where:

- $\mathcal{N}(A_1,.....,A_n)$ is The joint certainty distribution.

- min $(N(A_i|U(A_i))$ is the lower bound of the necessities degrees associated to $(A_i|U(A_i)$

*Example*: let the prior necessities and the conditional necessities be as described in table 1:

Table 1: necessities distribution

| a | 0.6 |
|---|---|
| ¬a | 0.1 |

| B\A | a | ¬a |
|---|---|---|
| b | 0.5 | 0.2 |
| ¬b | 0.25 | 0 |

| C\A | a | ¬a |
|---|---|---|
| c | 0.3 | 0.2 |
| ¬c | 0.1 | 0.1 |

| D\BC | bc | b¬c | Else |
|---|---|---|---|
| d | 0.2 | 0.1 | 0.3 |
| ¬d | 0.4 | 0.1 | 0.2 |

By the use of the chain rule defined by equation (13) we obtain the certain distribution associated with the certain Bayesian network cited above as described in table 2.

Table 2: joint necessity distribution

| A | B | C | D | *minN* |
|---|---|---|---|---|
| a | b | c | d | 0.2 |
| a | b | c | ¬d | 0.3 |
| a | b | ¬c | d | 0.1 |
| a | b | ¬c | ¬d | 0.3 |
| a | ¬b | c | d | 0.2 |
| a | ¬b | c | ¬d | 0.1 |
| a | ¬b | ¬c | d | 0.1 |
| a | ¬b | ¬c | ¬d | 0.1 |
| ¬a | b | c | d | 0.1 |
| ¬a | b | c | ¬d | 0.1 |
| ¬a | b | ¬c | d | 0.1 |
| ¬a | b | ¬c | ¬d | 0.1 |
| ¬a | ¬b | c | d | 0 |
| ¬a | ¬b | c | ¬d | 0 |
| ¬a | ¬b | ¬c | d | 0 |
| ¬a | ¬b | ¬c | ¬d | 0 |

## 4.2. From CBN to certain knowledge base
We would like to represent a class of certain Bayesian networks using a local certain valued knowledge base consisting of a collection of possibilistic logic sentences (formulae) in such a way that a network generated on the basis of the information contained in the knowledge base is isomorphic to a set of ground instances of the formulae. As the formal representation of the knowledge base, we use a set of certain formulae. We represent random variables with necessities weights. Formally a certain valued knowledge base is defined as the set:

$$\sum = \{(\varphi_i, \alpha_i), i = 1....m\} \qquad (14)$$

Where $\varphi_i$ denotes a classical propositional formula, and $\alpha_i$ denotes the lower bound of certainty (i.e necessity).

We can represent the information contained in each node of a Bayesian network, as well as the quantitative information contained in the link matrices, if we can represent all the direct parent/child relations. We express the relation between each random variable and its parents over a class of networks with a collection of quantified formulae. The collection of formulae represents the relation between the random variable and its parents for any ground instantiation of the quantified variables. The network fragment consisting of a random variable and its parents with a set of formulae of the form $(\varphi, \alpha)$.

We give next some definitions inspired from [9] and [10].

**Definition 1**:
*Two certain knowledge bases $\sum_1$ and and $\sum_2$ are said to be equivalent if their associated necessity distributions are equal, namely:*

$$\forall \omega \in \Omega, \ N\sum_1(\omega) = N\sum_2(\omega) \qquad (15)$$

**Definition 2**:
*Let $(\varphi, \alpha)$ be a formula in $\sum$ Then $(\varphi, \alpha)$ is said to be subsumed by $\sum$ if $\sum$ and $\sum\setminus\{(\varphi, \alpha)\}$ are equivalent knowledge bases.*

This is means that each redundant formula should be removed from the certain valued knowledge base since it can be deduced from the rest of formulae. Next, we describe the process that permit to deduce a certain valued knowledge base from a certain network.

Let *NG* be a certain Bayesian network consisting of a set of labeled variables $V = \{A_1, A_2...A_n\}$. Now let *A* be a binary variable and let $(a \ \neg a)$ be its instances. Given the measure $N(a_i|u_i)$ witch represents the local necessity degree associated with the variable *A* where $u_i \in U_A$ is an instance of parents$(a_i)$. The local certain knowledge base associated with *A* should be defined using the next equation:

$$\sum_A = \{(\neg a_i \vee u_i, \alpha_i), \alpha_i = 1 - N(a_i|u_i) \neq 0\} \qquad (16)$$

To note here that in [15] the authors prove the possibility to recover conditional possibilities from $\sum_A$ where $\sum_A$ is a possibilistic knowledge base.

Based o the results obtained in [9], we can check in our case that it is possible to recover conditional necessities from $\sum_A$ according to equations (17).

$$N\Sigma(\omega) = \begin{cases} 1 \text{ if } \forall \ (\varphi_i, \ \alpha_i) \in \Sigma \ \omega \models \varphi_i \\ 1\text{- max } \{ \ \alpha_i . \ \omega \ \not\models \varphi_i \} \text{ otherwise} \end{cases} \quad (17)$$

*Example*: by applying equation (16), we get the certain knowledge base associated to the certain Bayesian network described in section 4.1.

$\sum_A = \{(a, \ 0.9 \ )\}$

$\sum_B = \{(b \vee a, 0.7), (b \vee \neg a, 0.75)(\neg b \vee a, \ 0.8)\}$

$\sum_C = \{(c \vee a, \ 0.9), (c \vee \neg a, \ 0.9) \ (\neg c \vee a, \ 0.8)\}$

$\sum_D = \{\{(d \vee b \vee c, \ 0.8), \ (d \vee b \vee \neg c, \ 0.8), \ (d \vee \neg b \vee c, 0.9), (d \vee \neg b \vee \neg c, \ 0.6 \ ), \ (\neg d \vee \neg b \vee c, \ 0.9 \ )\}$

Next section shows the other form of certain valued knowledge where weighted formulae will be replaced by fuzzy membership function de define a new representation of uncertain information by the mean of what we call fuzzy knowledge bases.

## 5. CBN based on fuzzy necessity distribution

Logical formulae with a weight strictly greater than a given levels (lower bounds of necessity degrees) are immune to inconsistency and can be safely used in deductive reasoning [11]. However in order to perform reasoning for both imprecise and uncertain information, two important issues should be addressed. First, any improvement of the possibility level for a piece of information can only be achieved at the expense of the specificity of the information; second the accorded levels to the causality explained in terms of rules (case of fuzzy logic) and conditional dependencies (case of Bayesian networks) are somewhat expensive due to the fact that these confidence level is somewhat critical.

We propose so to combine these three approaches (Bayesian networks certainty distribution and fuzzy logic) to develop a method for uncertain and imprecise knowledge representation that may improve decision based systems.

Our fuzzy beliefs are to emulate a certain Bayesian necessity measure. For simplicity each variable here has two states: the presence or absence of an entity. The belief that $A$ is present takes the form of a fuzzy truth $f_A$. The extent to witch the belief of variable state influences the state beliefs of parent or child is modelled by a fuzzy set membership function: one for each influence direction.
*Example:*
Let our certain network be as described in figure representing a Bayesian network in metastatic cancer.

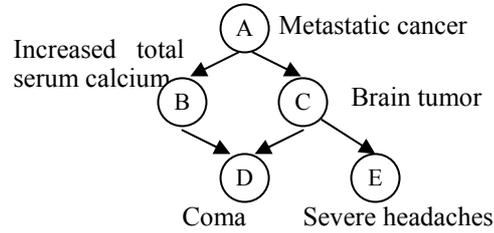

*Fig 4. A Bayesian network for metastatic cancer[12]*

Fig. 4 shows a Bayesian network representing the above cause and effect relationships. Table 2 lists the causal influences in terms of fuzzy certainty distributions. Each variable is characterized by an unknown necessity degree given the state of its parents. For instance: $C \in [0, 1]$ represents the dichotomy between having a brain tumor and not having one, c denotes the assertion C = 1 or "Brain tumor is present", and $\neg c$ is the negation of c, namely, $\bar{C}$ =0. The root node, A, which has no parent, is characterized by its prior fuzzy certainty distribution.

*Example*

Le the conditional fuzzy necessities associated to the graph presented in figure 4 be as described in table 3. For reason of simplicity we kept here four nodes only as in the graph presented in figure 3.

Table 3:  fuzzy necessity distribution

| a | $\neg a$ |
|---|---|
| [$\beta_{A11}$ , $\beta_{A12}$] | [$\beta_{A21}$ , $\beta_{A22}$] |

| B\A | a | $\neg a$ |
|---|---|---|
| b | [$\beta_{B|A11}$ , $\beta_{B|A12}$] | [$\beta_{B|A21}$ , $\beta_{B|A12}$] |
| $\neg b$ | [$\beta_{B|A31}$ , $\beta_{B|A32}$] | [$\beta_{B|A41}$ , $\beta_{B|A42}$] |

| C\A | a | $\neg a$ |
|---|---|---|
| c | [$\beta_{C|A11}$ , $\beta_{C|A12}$] | [$\beta_{C|A21}$ , $\beta_{C|A12}$] |
| $\neg c$ | [$\beta_{C|A31}$ , $\beta_{C|A32}$] | [$\beta_{C|A41}$ , $\beta_{C|A42}$] |

| D\BC | bc | b$\neg$c | Else |
|---|---|---|---|
| d | [$\beta_{D|BC11}$,$\beta_{D|BC12}$] | [$\beta_{D|BC21}$,$\beta_{D|BC22}$] | [$\beta_{D|BC31}$,$\beta_{D|BC32}$] |
| $\neg d$ | [$\beta_{D|BC41}$,$\beta_{D|BC42}$] | [$\beta_{D|BC51}$,$\beta_{D|BC52}$] | [$\beta_{D|BC61}$,$\beta_{D|BC62}$] |

For instance  N(d| b,$\neg$c) cannot be 0.1 as described in table 1 but rather is a fuzzy number  say $\chi_1$ $\in$[$\beta_{D|BC1}$ , $\beta_{D|BC2}$ ] where $\chi_1$= $\aleph$(d| b,$\neg$c) is the fuzzy necessity associated with the fuzzy formula (d| b,$\neg$c) and is associated with a membership function $\mu$ ( $\chi_1$ ) supposed to be a triangular function (respectively $\mu$ can be trapezoid or other kind of functions). $\mu$ is represented as follow (figure 5):

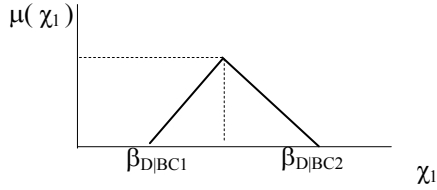

*Fig. 5:  a membership function*

Then we can deduce the next possible representation of $\mu(\chi_1)$ as:

$$\mu(\chi_1) = k_1 \text{ x } (\chi_1 - \beta_{D|BC1}) - k_2 \text{ x } (|\chi_1 - \beta_{D|BC2}| + \chi_1 - \alpha)$$

Where:

- $\alpha$, $k_1$ and $k_2$ are two defined constants..
- $| * |$ is the absolute  value of term $*$

The above expression and figure mean that the interval of $\chi_1$ is  $[\beta_{D|BC1}$ , $\beta_{D|BC2}]$. If $\chi_1 = \alpha$ then $\mu(\chi_1)=1$, implying that  the fuzzy necessity $\chi_1 = \alpha$ is the most possible situation. If $\chi_1 \geq \beta_{D|BC2}$ or $\chi_1 \leq \beta_{D|BC1}$     then     $\mu(\chi_1) = 0$, the possible manifestation of $\chi_1$.

## 6. Transformation between FBN and FNB

Analogously, when the given necessities degree are fuzzy numbers as we described in section 5, the necessity distribution $N(X)$ associated to a node X is considered as a fuzzy distribution defined by a membership function

$$\mu : [\beta_1, \beta_2] \longrightarrow [0\ 1] \qquad (18)$$
$$\chi \longmapsto \mu(\chi)$$

Example: consider the graph of figure 3. For simplicity each variable here has two states: the presence or absence of an entity and we will define the same membership function to $a$ as to $\neg a$.

Table 4:   fuzzy necessity distribution with membership functions

| a | ¬a |
|---|---|
| $[\beta_{A11}, \beta_{A12}]$ | $[\beta_{A21}, \beta_{A22}]$ |
| $\mu_1(\chi)$ | $\mu_1(\chi)$ |

| B\|A | a | ¬a |
|---|---|---|
| b | $[\beta_{B|A11}, \beta_{B|A12}]$ | $[\beta_{B|A21}, \beta_{B|A12}]$ |
| | $\mu_2(\chi)$ | $\mu_3(\chi)$ |
| ¬b | $[\beta_{B|A31}, \beta_{B|A32}]$ | $[\beta_{B|A41}, \beta_{B|A42}]$ |
| | $\mu_2(\chi)$ | $\mu_3(\chi)$ |

| C\|A | a | ¬a |
|---|---|---|
| c | $[\beta_{C|A11}, \beta_{C|A12}]$ | $[\beta_{C|A21}, \beta_{C|A12}]$ |
| | $\mu_4(\chi)$ | $\mu_5(\chi)$ |
| ¬c | $[\beta_{C|A31}, \beta_{C|A32}]$ | $[\beta_{C|A41}, \beta_{C|A42}]$ |
| | $\mu_4(\chi)$ | $\mu_5(\chi)$ |

| D\|BC | bc | b¬c | Else |
|---|---|---|---|
| d | $[\beta_{D|BC11}, \beta_{D|BC12}]$ | $[\beta_{D|BC21}, \beta_{D|BC22}]$ | $[\beta_{D|BC31}, \beta_{D|BC32}]$ |
| | $\mu_6(\chi)$ | $\mu_7(\chi)$ | $\mu_8(\chi)$ |
| ¬d | $[\beta_{D|BC41}, \beta_{D|BC42}]$ | $[\beta_{D|BC51}, \beta_{D|BC52}]$ | $[\beta_{D|BC61}, \beta_{D|BC62}]$ |
| | $\mu_1(\chi)$ | $\mu_7(\chi)$ | $\mu_8(\chi)$ |

Let the different membership be as follow:

$$\mu_i(\chi) = k_{i1} \text{ x } (\chi - \beta_{ij1}) - k_{i2} \text{ x } (|\chi - \beta_{ij2}| + \chi - \alpha_i)$$

Where:

- $\mu_i(\chi)$ is the membership function associated to the fuzzy variable $\chi$,  supposed to be triangular.
- $k_{i1}$ and $k_{i2}$ are the used constant in each membership function   supposed to be triangular.
- $\beta_{ij1}$ and $\beta_{ij2}$  are the two min and the max boundary of a necessity degree.

Finally by maximization of each membership function, we can deduce an optimal value for the certainty degree associated to each fuzzy variable (i.e. proposition). Namely:

$$\aleph(\chi) = \mu(\chi) = 1$$

Then it will be easy to deduce the value of $\chi$ as follow:

$$\chi = \frac{\lambda + k_{i1} \text{ x } \beta_{ij1} + k_{i2} \text{ x } \beta_{ij2} + 1}{k_{i1}} \qquad (19)$$

By replacing $\lambda$ by 1 (the maximization of $\mu(\chi)$), the value of $\chi$ will be:

$$\chi = \frac{\lambda + k_{i1} \text{ x } \beta_{ij1} + k_{i2} \text{ x } \beta_{ij2} + 1}{k_{i1}} \qquad (20)$$

Analogously, the definition of the fuzzy joint necessity distribution is obtained by applying the fuzzy chain rule:

$$\aleph(A_1,...,A_n) = \min(\chi_i), \chi_i = \aleph(A_i|U(A_i)$$

From a semantic point of view, a certain knowledge base $\Sigma = \{(\varphi_i, \alpha_i), i = 1....m\}$ where each $\alpha_i$ a crisp necessity value, is understood  as the necessity distribution $N_\Sigma$ representing the fuzzy sets of models of $\Sigma$ :

$N_\Sigma(\omega) = \min \max (\mu_{[Pi]}(\omega), 1-\alpha)$ where $[P_i]$ denotes the set of models of $P_i$, so that :

$$\mu_{[Pi]}(\omega) = \begin{cases} \mu_{[Pi]} = \alpha & \text{if } \omega \ P_i \\ 0 & \text{otherwise} \end{cases} \qquad (21)$$

From (21) we can clearly deduce clearly that $N_{\Sigma}(\omega)$ is naturally a fuzzy distribution applied to a crisp set of values and $\mu_{[Pi]}$ is the crisp membership function.

Finally a fuzzy knowledge base can be defined so as the set of fuzzy certain formulae. We represent random variables with fuzzy necessities weights. Formally a fuzzy necessity valued knowledge base $\Im$ is defined as the set:

$$\Im = \{(\varphi_i, \ \mu_i^{-1}(1)), \ i = 1\ldots.m\} \qquad (22)$$

Given that:

- $\varphi_i = (\neg a_i \vee u_i)$

- $\mu_i$ is the fuzzy membership function

- $\chi = 1 - \aleph \ (a_i|u_i) \ \neq 0$

- $\mu_i^{-1}(1) = \chi$ the reciprocal function of $\mu_i$

## 7. Conclusion and discussion

The present paper pretended to be an attempt to show and understand partially the world complexity that is being increasingly observed, being a try to contribute to ordering and organizing it. The considered approach combined the fuzzy logic and possibilistic logic in a framework of causal networks, which makes it an efficient tool that models knowledge taking into account the expert's subjectivity, vagueness and imprecision.

Certain Bayesian networks with fuzzy knowledge bases approach in a natural way gives us the subsethood of the evidence for each logical formula. Although the methodology proposed in this paper, is aimed and illustrated by some typical examples, the developed techniques require experimental results.

A future work is to extend this representation by definition of efficient algorithms for locally inferences. A future work will consist also on validating this approach by applying methodologies developed here on a real problem of knowledge representation with large scale data.